\documentclass[12pt,a4paper]{article}

\usepackage[
top=2.5cm,
bottom=2.5cm,
left=2.5cm,
right=2.5cm
]{geometry}

\usepackage[T1]{fontenc}
\usepackage[utf8]{inputenc}
\usepackage{lmodern}

\usepackage{longtable}
\usepackage{booktabs}
\usepackage{array}
\usepackage{longtable}
\usepackage{booktabs}
\usepackage{array}
\usepackage{tipa}

\DeclareUnicodeCharacter{2032}{\ensuremath{^\prime}}

\usepackage{amsmath}
\usepackage{amssymb}
\usepackage{amsthm}

\usepackage{booktabs}
\usepackage{array}
\usepackage{longtable}
\usepackage{multirow}
\usepackage{float}

\usepackage{graphicx}
\usepackage{caption}
\usepackage{subcaption}

\usepackage{enumitem}

\usepackage{setspace}

\usepackage{microtype}

\usepackage[
colorlinks=true,
linkcolor=blue,
citecolor=blue,
urlcolor=blue
]{hyperref}

\usepackage{amsmath}
\usepackage{tabularx}
\usepackage{booktabs}
\usepackage{float}

\usepackage[
backend=biber,
style=vancouver,
sorting=none
]{biblatex}

\numberwithin{equation}{section}

\title{
	\textbf{Using Letter Positional Probabilities to Assess French Word Complexity}
}

\author{
	Michael Dalvean\\
	\small Albany, Western Australia\\
	\small October 2026\\
	\small \texttt{admin@michaeldalvean.com}
}

\date{}

\begin{document}
	
	\maketitle

	
	\begin{abstract}
		
		We use Letter Positional Probabilities (LPPs) to examine the orthographic differences between simple and complex French words. We draw from a sample of 38,000 French words scored by reaction time on a lexical decision task. We take the 5000 highest reaction time words and the 5000 lowest as a proxy for high and low word complexity groups respectively. We create 18 variables based on the structure of the words, such as length and number of syllables. We also reduce each word to a string of binary values by one-hot encoding each letter according to the 42 characters of the standard French alphabet up to a length of 25 letters. Thus, each word is represented by 42 × 25 binary characters. We then create LPPs by looking at the average binary scores for simple as opposed to complex words for all letter/positions. We find that there is a high number of LPPs that are associated with word complexity. For example, we find that most accented letters are associated with simpler words. We examine the effects of structural and LPP variables in classifiers and find that word length alone enables a classification accuracy of 77.27\%, while adding the binary variables from the first 6 letter positions provided an accuracy of 83.95\%. We discuss possible explanations as to how complexity could be encoded in orthography.

	\end{abstract}


	\section{Introduction}
	
	This paper examines French word complexity using letter positional probabilities (LPPs). In \cite{dalvean2024using} it was demonstrated across three different datasets that LPPs can be used to assess English word complexity. In this study we apply the same concept to French word complexity. 
	The underlying hypothesis here is that there is a latent quality of words which we call word complexity. This quality can be approximated with a number of proxies such as word length, number of syllables, age of acquisition, frequency, concreteness, orthographic characteristics and reaction times on various lexical decision criteria \cite{Ferrand2011Chronolex,Balota2007EnglishLexicon,YapBalota2009Multisyllabic,Brysbaert2016Prevalence,BalotaChumbley1984LexicalAccess,RaynerDuffy1986LexicalComplexity,Keuleers2012BritishLexicon}. The idea of using LPPs is that these may also be proxies for word complexity. 
	
	We use the reaction time on a lexical decision task as a proxy for word complexity to select a sample of 5000 short reaction time (low complexity) words and 5000 long reaction time (high complexity) words from a database of over 38000 French words. Thus, the samples are drawn from the extreme ends of the frequency distribution. In this way, we approximate a sample that represents the extreme ends of the complexity range.
	
	An important caveat that we must consider is the extent to which reaction times are a proxy for complexity. It is certainly established that reaction times are associated with the usual proxies noted above. Given that these are used as proxies for word complexity, it seems reasonable to claim that reaction times tap into significant elements of word complexity.
	
	We generate LPPs by `one-hot encoding', for each letter position, a binary for 1 - 26 Latin characters plus 16 accented letters and ligatures. We do this for word lengths of 3 - 25 letters.  The purpose of this is to represent the fine grain 'atomic' structure of the word. We find that there are many LPPs that are highly statistically significant in distinguishing between high and low complexity words.

	\section{Previous Work}
	\label{sec:literature}
	
	This study extends the approach  in \cite{dalvean2024using} where LPPs were used to assess English word complexity over three datasets. In the first two datasets, 66 LPPS were found to be common as between the two datasets (p < .001), and these were then used to create a classifier that was able to correctly classify a third dataset with an accuracy of 70\%. Thus, LPPs seem to carry information relating to word complexity. However, these experiments used English words. The current study attempts to find a similar phenomenon with French words.
	
	The approach that is most similar to the current study is that exemplified by \cite{vitevitch1999probabilistic}, in which phonotactic segments of words were used rather than individual letters. The idea here is that phonotactic segments that are more common in a language are more easily recognised. In that study, it was shown that phonotactic segments that are common in a language are more easily recognised in spoken words than those that are not, and this facilitates word recognition. However, there is an opposing influence in that, if there is a high density of words that have similar phonotactic segments, recognition time can slow due to the effort required to disambiguate the target word from other competing words.
	
	Phonological segment recognition has been found to occur in infants as young as 9 months\cite{jusczyk1993infants}. In that study, infants were found to be able to recognise phonotactic segments from their own language. Similarly, native speakers of Dutch use phonotactic segments to separate individual words in fluent speech \cite{mcqueen1998segmentation}. There is evidence that phonotactic probability also affects word production. In \cite{vitevitch2004sublexical}, subjects named pictures with more probable phonotactic segments  words more quickly than words with lower phonotactic probability. However, those target words with high numbers of orthographical neighbours where the initial phoneme is shared by the target word (high onset density), were responded to more slowly than target words with fewer  orthographic neighbours sharing an initial phoneme (low onset density). Finally, in \cite{Frisch2000Wordlikeness}, those nonwords that contained high probability phonological segments were judged to be more 'wordlike' than nonwords made of low-probability segments.

	In \cite{soler2018comparative} an algorithm was designed to score synonyms by simplicity on a continuous scale. The training took place with two sources with graded levels of vocabulary for younger/older students and French as a foreign language students. It was evaluated by comparing the rankings of synonyms by simplicity coded manually. They used 300-dimension word embeddings as features. Here the embeddings were vector representations of the word environment in which a target word appears in a large training corpus. They combined word embeddings combined with number of characters, number of phonemes and log frequency. In isolation, embeddings perform better than other features, but the best set of predictors was number of characters and + log frequency.
	
	In \cite{gala2014model}, using the same graded corpora as used in \cite{soler2018comparative}, the authors calculated nine structural variables such as number of characters, number of syllables, phonemes etc. They also used orthographic neighbourhood, phoneme-grapheme correspondence - based on a three point scale. The second set of features was based on morphology, such as number of morphemes, suffixation, prefixation etc. A third set of features consisted of two measures of polysemy. Finally, they include two measures of word frequency. In both corpora, frequency was the feature with the highest correlation with complexity. The overall finding was that word frequency provided more signal than the other variables and the addition of other variables provided only marginal improvement. They found that intermediate classes are the most difficult to classify, with the extremes more easy. The authors acknowledged that the labels may themselves have been noisy, given that the same word may be inlcuded in or excluded from different corpora based on textbook selection and teaching approaches. They also point out that the morphology data may have been inaccurate.
	
	In \cite{aleksandrova2023cefr} the authors used word embeddings in order to capture word meanings. The idea, as in other work using embeddings, is that a word is given its function in relation to other words, the difference being that they include a context for each word in their word embedding procedure. Thus, the word `run' has 10 entries, and this includes two noun
	forms and eight verbs. The lowest level of language learner (A1) learns the word in the context "He can run very fast" but at C2, "He would like to run for mayor". The procedure used involved presenting each target word with a contextual sentence. Contextual embeddings improved the Macro-F1 score across A1 - C2 language learning levels by 22 percentage points in comparison to using frequency and word length alone. An attempt to use multilingual BERT to score a corpus of French words using the model trained on English words was less successful, with only the beginner word class being correctly scored. The authors mentioned possible reasons for this as being differences in morphology between English and French and the lack of availability of French lexical sources.

	Alfter et al. \cite{alfter2022dictionary} examined word-sense difficulty as perceived among French native and non-native speakers. Participants ranked word senses as presented in individual sentences from easiest to most difficult. There was a high degree of correlation between the native and non-native French speakers. Furthermore, the word sense difficulty rankings were broadly correlated with vocabulary difficulty established in language-learning environments. A subsequent attempt to find an association between word-sense difficulty and the complexity of dictionary definitions was inconclusive.

In \cite{grainger1993masked}, subjects were given a prime and were subsequently given a lexical decision task. The format of the primes, for a given target ABCDE, was \%BCDE, A\%CDE, AB\%DE, ABC\%E and ABCD\%E. They found that some partial word primes were more effective than others at eliciting the correct response. The important factor was the letter positional frequency. Where a given letter is unlikely to appear in a given position in words in the language in general, there is less room to move in terms of the decision as to the correct target, and so the prime is more efficient.

In \cite{mahowald2018word}, it was found that more frequent words are `better formed'. A word is well formed if its sequence of letters, based on a trigram search algorithm, is probable given other sequences in words in the language. They also found that more frequent words had more orthographic neighbours.

	In \cite{carus2016new}, a compression algorithm (CAFTS) designed to speed up text search was used to compress texts from various languages to approximately 41\% of their original size. The algorithm is based on a series of dictionaries of trigrams. Using the algorithm, French had a compression ratio of 3.18 bits/character, which was lower on average than several other languages,  with an average across 8 languages of 3.31. We reproduce the table below as Table~\ref{tab:compression-ratios} as it includes the compression data for several languages and compression ratios.

\begin{table}[H]
	\centering
	\caption{Compression ratios (in bits/character), compression and
		decompression times (in seconds).}
	\label{tab:compression-ratios}
	
	\begin{tabular}{p{1.5cm}rrrrrr}
		\hline
		& \textbf{Gzip} & \textbf{Compress} & \textbf{DLETDC} &
		\textbf{ETDC} & \textbf{SCDC} & \textbf{CAFTS} \\
		\hline
		Dutch   & 2.86 & 3.04 & 2.91 & 2.84 & 2.80 & 3.23 \\
		English & 3.06 & 3.22 & 2.82 & 2.78 & 2.70 & 3.23 \\
		Finnish & 3.03 & 3.25 & 4.12 & 4.02 & 3.99 & 3.36 \\
		French  & 2.87 & 3.11 & 3.15 & 3.09 & 3.04 & 3.18 \\
		German  & 2.98 & 3.24 & 3.40 & 3.32 & 3.28 & 3.45 \\
		Italian & 3.10 & 3.24 & 3.29 & 3.21 & 3.17 & 3.29 \\
		Spanish & 2.93 & 3.15 & 3.33 & 3.26 & 3.21 & 3.32 \\
		Turkish & 3.09 & 3.43 & 4.16 & 4.08 & 4.01 & 3.44 \\
		\hline
		Average ratio & 2.99 & 3.21 & 3.40 & 3.33 & 3.28 & 3.31 \\
		Comp' time    & 20.65 & 5.15 & 4.83 & 6.24 & 6.32 & 36.20 \\
		Decomp' time  & 1.66 & 1.55 & 1.95 & 1.94 & 1.91 & 1.26 \\
		\hline
	\end{tabular}
	
	\smallskip
	\begin{minipage}{0.85\textwidth}
		\footnotesize
		\textit{Source:} Carus and Mesut \cite[p.~4363]{carus2016new}.
	\end{minipage}
\end{table}

\par\smallskip

	\section{Data and Methods}
	\label{sec:methods}
	
	\subsection{Data}
	
	The data for this study comes from \cite{ferrand2010french}. This source contains 38,840 French words and 38,840 non-words. The study was designed to determine the time taken to decide whether a given word was a real word or a non word. We used the z scores for the reaction times to rank the words.

	\subsection{Data Cleaning}
	
	With 38,840 words, we undertake the following data cleaning steps.
	
	\begin{itemize}
		\item convert characters to lowercase. The original words were in lowercase, but this step was undertaken to ensure there were no capitals
		\item remove all words containing any character other than those of the 42-character standard French alphabet
		\item remove lemmas with non-alphabetical characters
		\item  remove words shorter than 3 characters or longer than 25 characters

	\end{itemize}
	
	After these operations we are left with 38,287 words. Of these, we extract the 5000 with the lowest reaction times and the 5000 with the highest reaction times, giving us a sample of n = 10000.
	
	\subsection{Variables}
	
	 The fundamental analysis in this study is based on LPPs. The method of calculating these is as follows.
	 
	 For any word of up to 25 letters, each letter in positions 1...25 is represented by 42 binary variables, each of which is a binary representing the presence/absence of one of the 42 letters of the alphabet. Thus, we have 25 positions × 42 alphabetic variables. For each position, if a given letter of the alphabet is in that position, the binary for that letter takes the value of 1 for that letter and 0 for all other 41 letters in that position. For example, for the word 'cat', with 'c' in the first position (the first 42 variables) the letter 'c' takes a binary value of 1 and the other letters take  a 0. For position two (variables 43 - 84), the letter 'a' takes a binary value of 1, and the other letters take a value of 0. Finally, for position 3 (variables 85 - 126), the 't' takes a binary value of one and all other letters take a value of 0. For the remaining letter positions from 4 - 25, all values = 0. 
	 
Thus, for any word \textit{w}
\[
w=(w_1,w_2,\ldots,w_L),
\]
given \(3 \leq L \leq 25\), each character \(w_i\) belongs
to the set of French alphabet characters
\[
\mathcal{A}
=
\{a,b,c,d,e,f,g,h,i,j,k,l,m,n,o,p,q,r,s,t,u,v,w,x,y,z,
\text{à},\text{â},\text{æ},\text{ç},\text{é},\text{è},
\text{ê},\text{ë},\text{î},\text{ï},\text{ô},\text{œ},
\text{ù},\text{û},\text{ü},\text{ÿ}\},
\]
and \(|\mathcal{A}|=42\).

We define the binary positional variable as
\[
x_{ij}(w)=
\begin{cases}
	1, & \text{if } i \leq L \text{ and } w_i=a_j,\\[4pt]
	0, & \text{otherwise},
\end{cases}
\]
where
\[
i=1,\ldots,25
\]
is the character position in the word, and
\[
j=1,\ldots,42
\]
is the index of alphabet characters.

The unaccented characters in  \(\mathcal{A}\) above are represented by the first 26 indices:
 
\[
a_1=a,\quad a_2=b,\quad \ldots,\quad a_{26}=z.
\]

The accented characters and ligatures in \(\mathcal{A}\) above are represented by the final 16 indices.
characters and ligatures:
\[
\begin{aligned}
	a_{27}&=\text{à}, &
	a_{28}&=\text{â}, &
	a_{29}&=\text{æ}, &
	a_{30}&=\text{ç},\\
	a_{31}&=\text{é}, &
	a_{32}&=\text{è}, &
	a_{33}&=\text{ê}, &
	a_{34}&=\text{ë},\\
	a_{35}&=\text{î}, &
	a_{36}&=\text{ï}, &
	a_{37}&=\text{ô}, &
	a_{38}&=\text{œ},\\
	a_{39}&=\text{ù}, &
	a_{40}&=\text{û}, &
	a_{41}&=\text{ü}, &
	a_{42}&=\text{ÿ}.
\end{aligned}
\]

Given that there are 25 possible positions and
42 possible characters at each position, the
complete set of binary variables for each word contains
\[
25 \times 42 = 1050
\]
positional indicators.

For words of length < 25, the binary values beyond the position of its final letter are given a value of 0. It should be noted that, in the modeling stage, any binary variable that has zero values for all words is removed from the analysis. In this way we are able to reduce the number of variables from a potential 1050 to 459 for the corpus under study. 

The above describes the creation of binaries for each letter position. These binary values are used in the creation of classifiers, where the target is a 1 for complex words and 0 for simple words. The LPPs are calculated by averaging the binaries for each of the simple and complex groups. This gives the probabilities that a given letter will occur at a given position for a simple as opposed to a complex word.

Given a group of words selected for their simplicity/complexity \(G\), the LPP for the alphabet character \(a_j\) at the position \(i\)
is the average of the binary indicators for the same alphabet character at the same position for all words in that group:
\[
P_{ij}^{(G)}
=
\frac{1}{N_G}
\sum_{w \in G} x_{ij}(w),
\]
where \(N_G\) is the number of words in group \(G\).

The group means calculated in this way are used for the two tailed t-tests in the descriptive statistical analysis below.

For each of the two groups, we calculate the difference in positional probability for a given character at a given position as

\[
\Delta P_{ij}
=
P_{ij}^{(\mathrm{high})}
-
P_{ij}^{(\mathrm{low})}.
\]

Where \(\Delta P_{ij}\) is positive, the \(a_j\) character occurs more often at position \(i\) in complex words, and a negative value indicates more common occurrence in simple words.

	We also calculate 18 variables based on the structure of the words that are associated with word complexity. 
	
	\begin{itemize}
		
		\item LENGTH = n; measures basic orthographic size.
		
		\item SYLLABLES = number of syllables in the word. Note that the number of syllables is not calculated  from the Python script because the method of calculating syllables from French words was not accurate. The number of syllables was therefore taken from the original data source (\cite{ferrand2010french}). 
		
		\item MEAN\_SYLLABLE\_LENGTH; measures average syllable size.
		
		\item UNIQUE\_LETTERS; counts the number of different letters occurring in the word.
		
		\item LETTER\_DIVERSITY; measures normalized character diversity.
		
		\item REPEATED\_LETTERS; measures repetition/redundancy of characters.
		
		\item DOUBLE\_LETTERS; counts adjacent identical letter pairs.
		
		\item VOWELS; counts vowels in the word.
		
		\item CONSONANTS; number of alphabetic characters that are not vowels.
		
		\item VOWEL\_CONSONANT\_RATIO; measures Vowel/Consonant composition
		
		\item CONSONANT\_CLUSTERS; measures number of uninterrupted consonant sequences of length >= 2.
		
		\item MAX\_CONSONANT\_CLUSTER; maximum length of any uninterrupted consonant sequence.
		
		\item CV\_TRANSITIONS; counts changes between vowel and consonant categories.
		
		\item BIGRAM\_DIVERSITY; measures adjacent letter-pair diversity.
		
		\item TRIGRAM\_DIVERSITY; measures three-letter-sequence diversity.
		
		\item LETTER\_ENTROPY; measures uncertainty/diversity of letters within the word.
		
		\item REPETITION\_RATIO; measures internal redundancy.
		
		\item ORTH\_NEIGHBOURS; number of same-length words differing in exactly one letter.
		
	\end{itemize}

	With 10,000 words, we create a matrix of 1050 binary variables and 18 continuous variables. Of the 1050 binary variables, some have empty columns. Removing all empty columns gives us 441 LPP variables. Thus, the total number of variables is 459. 
		
	The target column consists of binary values, with 0 being assigned to the low complexity group and 1 to the high complexity group.

	There is a large number of independent variables under analysis. In subsequent analysis, we therefore make a Bonferroni correction to all p-values by multiplying them by 459.

	\section{Results}
	\label{sec:results}
	 After Bonferroni correction, there are 163 variables that are significant at the p<.01 level. These are reproduced at \ref{app:high-low-frequency}. Inspection of the table shows that  several of the standard word complexity measures are significantly discriminative of the high and low frequency groups, with the top 12 most significant variables being complexity measures. Of the 163 significant variables, 145 are LPPs, and all structural variables were found to be significant.
	 
	 In order to contextualise the results for the LPPs, we should consider that the average word length of the low complexity group is 6.69 and that of the high complexity group is 9.77. What this means is that letter positions above 6 are more likely to be significant purely because their presence is indicative of greater length, and length is highly associated with complexity. This is why the first 27 significant LPPs are from positions >6. We will examine the effects of this below when we create classifiers from the LPP data. At this stage, we will examine what is happening in the first 6 positions because these are more properly associated with orthography than length. 
	 
	 The significant LPPs  in positions < 7 are presented in Table \ref{tab:french-positional-features}, in which there are 42 significant variables from positions <7. 
	
	\begin{table}[H]
		\centering
		\caption{Low- and high-complexity French words: mean values for positional letter features at positions <7.}
		\label{tab:french-positional-features}
		
		\scriptsize
		\setlength{\tabcolsep}{4pt}
		\renewcommand{\arraystretch}{0.88}
		
		\begin{tabular}{@{}lccc@{}}
			\toprule
			\textbf{Variable}
			&
			\shortstack{\textbf{Mean -- Low}\\\textbf{Complexity}\\\textbf{($n=5{,}000$)}}
			&
			\shortstack{\textbf{Mean -- High}\\\textbf{Complexity}\\\textbf{($n=5{,}000$)}}
			&
			\shortstack{\textbf{$p$ value}\\\textbf{(Bonf. $\times459$)}}
			\\
			\midrule
			
			o6 & 0.011 & 0.051 & $<.0001$ \\
			e5 & 0.168 & 0.092 & $<.0001$ \\
			n2 & 0.035 & 0.086 & $<.0001$ \\
			a6 & 0.042 & 0.094 & $<.0001$ \\
			é6 & 0.056 & 0.017 & $<.0001$ \\
			i1 & 0.020 & 0.060 & $<.0001$ \\
			i6 & 0.053 & 0.105 & $<.0001$ \\
			e6 & 0.161 & 0.097 & $<.0001$ \\
			o2 & 0.195 & 0.129 & $<.0001$ \\
			o5 & 0.040 & 0.078 & $<.0001$ \\
			p6 & 0.004 & 0.020 & $<.0001$ \\
			j1 & 0.019 & 0.005 & $<.0001$ \\
			v1 & 0.044 & 0.021 & $<.0001$ \\
			l6 & 0.028 & 0.052 & $<.0001$ \\
			i5 & 0.089 & 0.127 & $<.0001$ \\
			z6 & 0.011 & 0.002 & $<.0001$ \\
			l2 & 0.043 & 0.022 & $<.0001$ \\
			a4 & 0.069 & 0.102 & $<.0001$ \\
			m6 & 0.014 & 0.031 & $<.0001$ \\
			ê2 & 0.007 & 0.000 & $<.0001$ \\
			m2 & 0.015 & 0.032 & $<.0001$ \\
			m5 & 0.012 & 0.028 & $<.0001$ \\
			r5 & 0.077 & 0.108 & $<.0001$ \\
			y3 & 0.014 & 0.004 & $<.0001$ \\
			c6 & 0.018 & 0.034 & $<.0001$ \\
			z5 & 0.007 & 0.001 & $<.0001$ \\
			f6 & 0.003 & 0.012 & $1.68\times10^{-4}$ \\
			u3 & 0.087 & 0.060 & $1.88\times10^{-4}$ \\
			r4 & 0.061 & 0.088 & $2.22\times10^{-4}$ \\
			â2 & 0.008 & 0.001 & $2.44\times10^{-4}$ \\
			l1 & 0.032 & 0.017 & $2.59\times10^{-4}$ \\
			é5 & 0.040 & 0.023 & $7.14\times10^{-4}$ \\
			h3 & 0.005 & 0.014 & $1.08\times10^{-3}$ \\
			a1 & 0.063 & 0.088 & $1.47\times10^{-3}$ \\
			t3 & 0.061 & 0.085 & $1.48\times10^{-3}$ \\
			o4 & 0.060 & 0.084 & $1.48\times10^{-3}$ \\
			û3 & 0.005 & 0.000 & $2.54\times10^{-3}$ \\
			g6 & 0.014 & 0.027 & $2.72\times10^{-3}$ \\
			f1 & 0.053 & 0.034 & $2.88\times10^{-3}$ \\
			r3 & 0.082 & 0.108 & $3.01\times10^{-3}$ \\
			t6 & 0.059 & 0.082 & $3.70\times10^{-3}$ \\
			x3 & 0.008 & 0.002 & $4.11\times10^{-3}$ \\
			â6 & 0.000 & 0.004 & $5.85\times10^{-3}$ \\
			
			\bottomrule
		\end{tabular}
	\end{table}
	 
	 One interesting observation is that accented vowels, other than â6, are associated with low complexity.
	 
	 \begin{table}[H]
	 	\centering
	 	\caption{Low- and high-complexity French words: Significant accented-character features at positions < 7}
	 	\label{tab:french-accented-characters}
	 	
	 	\begin{tabular}{@{}lccc@{}}
	 		\toprule
	 		\textbf{Feature} &
	 		\textbf{Low Mean} &
	 		\textbf{High Mean} &
	 		\textbf{Direction} \\
	 		\midrule
	 		
	 		é6 & 0.056 & 0.017 & \textbf{Low $\uparrow$} \\
	 		ê2 & 0.007 & 0.000 & \textbf{Low $\uparrow$} \\
	 		â2 & 0.008 & 0.001 & \textbf{Low $\uparrow$} \\
	 		é5 & 0.040 & 0.023 & \textbf{Low $\uparrow$} \\
	 		û3 & 0.005 & 0.000 & \textbf{Low $\uparrow$} \\
	 		â6 & 0.000 & 0.004 & High $\uparrow$ \\
	 		
	 		\bottomrule
	 	\end{tabular}
	 \end{table}

	On the other hand, the six significant unaccented vowels are associated with high complexity.

	\begin{table}[ht]
		\centering
		\caption{Low- and high-complexity French words: Significant unaccented-character features at positions < 7}
		\label{tab:french-unaccented-vowels}
		
		\begin{tabular}{@{}lccc@{}}
			\toprule
			\textbf{Feature} &
			\textbf{Low Mean} &
			\textbf{High Mean} &
			\textbf{Direction} \\
			\midrule
			
			a4 & 0.069 & 0.102 & High $\uparrow$ \\
			a6 & 0.042 & 0.094 & High $\uparrow$ \\
			i5 & 0.089 & 0.127 & High $\uparrow$ \\
			i6 & 0.053 & 0.105 & High $\uparrow$ \\
			o5 & 0.040 & 0.078 & High $\uparrow$ \\
			o6 & 0.011 & 0.051 & High $\uparrow$ \\
			
			\bottomrule
		\end{tabular}
	\end{table}

	In \cite{mathey2006orthographic} it was found that accents do not significantly affect lexical processing, so it is an open question why the above associations might exist, and it is beyond the scope of the current study to examine possible reasons. However, these exploratory findings indicate that there is information encoded in orthography that carries a simplicity/complexity signal.

	Position 6 is the most highly represented out of the first 6 positions: there are 6 vowels; o6, a6, é6, i6, e6, â6, and 8 consonants; p6, l6, z6, m6, c6, f6, g6, t6. There are four vowels, o, a, â6 and i, that are associated with high complexity words, while é6,and e6 are associated with low complexity words.

\begin{table}[H]
	\centering
	\caption{Low- and high-complexity French words: Significant vowel and consonant features at position 6.}
	\label{tab:french-position6}
	
	\begin{tabular}{@{}lccc@{}}
		\toprule
		\textbf{Feature} &
		\textbf{Low Mean} &
		\textbf{High Mean} &
		\textbf{Direction} \\
		\midrule
		
		\multicolumn{4}{@{}l}{\textit{Vowels}} \\
		\addlinespace[2pt]
		o6  & 0.011 & 0.051 & High $\uparrow$ \\
		a6  & 0.042 & 0.094 & High $\uparrow$ \\
		i6  & 0.053 & 0.105 & High $\uparrow$ \\
		â6  & 0.000 & 0.004 & High $\uparrow$ \\
		e6  & 0.161 & 0.097 & \textbf{Low $\uparrow$} \\
		é6  & 0.056 & 0.017 & \textbf{Low $\uparrow$} \\
		
		\addlinespace[4pt]
		\multicolumn{4}{@{}l}{\textit{Consonants}} \\
		\addlinespace[2pt]
		p6 & 0.004 & 0.020 & High $\uparrow$ \\
		l6 & 0.028 & 0.052 & High $\uparrow$ \\
		m6 & 0.014 & 0.031 & High $\uparrow$ \\
		c6 & 0.018 & 0.034 & High $\uparrow$ \\
		f6 & 0.003 & 0.012 & High $\uparrow$ \\
		g6 & 0.014 & 0.027 & High $\uparrow$ \\
		t6 & 0.059 & 0.082 & High $\uparrow$ \\
		z6 & 0.011 & 0.002 & \textbf{Low $\uparrow$} \\
		
		\bottomrule
	\end{tabular}
\end{table}	 

\noindent
	Seven of the eight consonants at position 6 are associated with high complexity, with z the exception.  
	 
At position 1, there is a distinct division of vowel and consonant associations, with the vowels a and i being associated with complex words and the consonants j,l, v and f associated with low complexity words.

	\begin{table}[H]
		\centering
		\caption{Low- and high-complexity French words: Significant character features at position 1.}
		\label{tab:french-position1-characters}
		
		\begin{tabular}{@{}lccc@{}}
			\toprule
			\textbf{Feature} &
			\textbf{Low Mean} &
			\textbf{High Mean} &
			\textbf{Direction} \\
			\midrule
			
			i1 & 0.020 & 0.060 & High $\uparrow$ \\
			j1 & 0.019 & 0.005 & \textbf{Low $\uparrow$} \\
			v1 & 0.044 & 0.021 & \textbf{Low $\uparrow$} \\
			l1 & 0.032 & 0.017 & \textbf{Low $\uparrow$} \\
			a1 & 0.063 & 0.088 & High $\uparrow$ \\
			f1 & 0.053 & 0.034 & \textbf{Low $\uparrow$} \\
			
			\bottomrule
		\end{tabular}
	\end{table}

The significant features at position 2 are as  depicted in Table~\ref{tab:french-position2-characters}

\begin{table}[H]
	\centering
	\caption{Low- and high-complexity French words: Significant character features at position 2.}
	\label{tab:french-position2-characters}
	
	\begin{tabular}{@{}lccc@{}}
		\toprule
		\textbf{Feature} &
		\textbf{Low Mean} &
		\textbf{High Mean} &
		\textbf{Direction} \\
		\midrule
		
		n2 & 0.035 & 0.086 & High $\uparrow$ \\
		o2 & 0.195 & 0.129 & \textbf{Low $\uparrow$} \\
		l2 & 0.043 & 0.022 & \textbf{Low $\uparrow$} \\
		ê2 & 0.007 & 0.000 & \textbf{Low $\uparrow$} \\
		m2 & 0.015 & 0.032 & High $\uparrow$ \\
		â2 & 0.008 & 0.001 & \textbf{Low $\uparrow$} \\
		
		\bottomrule
	\end{tabular}
\end{table}

We note that, as in the tables above, n and m are associated with high complexity. This occurs for the full cohort of results with m and n being associated with complexity for all significant occurrences of these letters in all positions.

	At position 3, high complexity is associated with the consonants h, t and r, while low complexity is associated with the vowels u, û, y,  and the only consonant associated with low complexity is x.

	\begin{table}[H]
		\centering
		\caption{Low- and high-complexity French words: Significant character features at position 3}
		\label{tab:french-position3-characters}
		
		\begin{tabular}{@{}lccc@{}}
			\toprule
			\textbf{Feature} &
			\textbf{Low Mean} &
			\textbf{High Mean} &
			\textbf{Direction} \\
			\midrule
			
			y3  & 0.014 & 0.004 & \textbf{Low $\uparrow$} \\
			u3  & 0.087 & 0.060 & \textbf{Low $\uparrow$} \\
			h3  & 0.005 & 0.014 & High $\uparrow$ \\
			t3  & 0.061 & 0.085 & High $\uparrow$ \\
			û3  & 0.005 & 0.000 & \textbf{Low $\uparrow$} \\
			r3  & 0.082 & 0.108 & High $\uparrow$ \\
			x3  & 0.008 & 0.002 & \textbf{Low $\uparrow$} \\
			
			\bottomrule
		\end{tabular}
	\end{table}

Position 4 has two vowels, a and o, and one consonant, r, all associated with high complexity.

\begin{table}[H]
	\centering
	\caption{Low- and high-complexity French words: Significant character features at position 3}
	\label{tab:french-position4-characters}
	
	\begin{tabular}{@{}lccc@{}}
		\toprule
		\textbf{Feature} &
		\textbf{Low Mean} &
		\textbf{High Mean} &
		\textbf{Direction} \\
		\midrule
		
		a4 & 0.069 & 0.102 & High $\uparrow$ \\
		r4 & 0.061 & 0.088 & High $\uparrow$ \\
		o4 & 0.060 & 0.084 & High $\uparrow$ \\
		
		\bottomrule
	\end{tabular}
\end{table}

	At position 5, there is no apparent vowel/consonant pattern, however it is worth noting that e and é both move in the same (low complexity) direction, as we noted above in position 6, indicating that both e and é in positions <7 are associated with low complexity.

	\begin{table}[H]
		\centering
		\caption{Low- and high-complexity French words: Significant character features at position 5}
		\label{tab:french-position5-characters}
		
		\begin{tabular}{@{}lccc@{}}
			\toprule
			\textbf{Feature} &
			\textbf{Low Mean} &
			\textbf{High Mean} &
			\textbf{Direction} \\
			\midrule
			
			e5  & 0.168 & 0.092 & \textbf{Low $\uparrow$} \\
			o5  & 0.040 & 0.078 & High $\uparrow$ \\
			i5  & 0.089 & 0.127 & High $\uparrow$ \\
			m5  & 0.012 & 0.028 & High $\uparrow$ \\
			r5  & 0.077 & 0.108 & High $\uparrow$ \\
			z5  & 0.007 & 0.001 & \textbf{Low $\uparrow$} \\
			é5  & 0.040 & 0.023 & \textbf{Low $\uparrow$} \\
			
			\bottomrule
		\end{tabular}
	\end{table}

In \cite{new2008differential} it was found that that consonants play a more important role in English word recognition during reading than vowels. In \cite{carreiras2009consonants} the study design involved target words with either consonants in first, thrd and fifth place (e.g., acero), or vowels in the same positions (e.g., farol). Primes consisted of the actual letters in those positions (a,c,o/f,r,l), voewls/consonants that did not appear in the word (l,u,i/t,s,b), or by the actual word (acero/farol). Using measurements of Event Related Potentials, it was found that the consonant primes had a similar Event Related Potentials effects to the actual word, while vowel primes had effects similar to the unrelated primes. Thus, there is some evidence that lexical processing may be different for the vowels as opposed to consonants. Brain activation has been found to be different in relation to vowel and consonant in several contexts \cite{carreiras2008brain}. Again, the evidence from the research is that there are possible reasons why vowels and consonants are differentially associated with complexity. However, there is no direct evidence of a link between vowels/consonants and word complexity.

	 \section {Classifiers}
	 
	 The results above indicate that there are significant variables that are likely to be able to be used to create classifiers to distinguish between low and high complexity words. By introducing different sets of variables, we can determine which variables carry the greatest signal. On the basis of this, we can determine whether the signals in the data described above are merely statistical phenomena, or whether there is something about the orthography of French words that carries information about word complexity.  
	 
	 The features used for the classifiers are the variables as described above. Thus, we have 441 LPP features, 18 structural features, and the target is a binary with 1 representing the high complexity words and 0 representing the low complexity words. Details of the classifiers are presented in Table \ref{tab:classifier-specifications} 
	 
\begin{table}[H]
	\centering
	\caption{Details of SVM, RF and NN classifiers}
	\label{tab:classifier-specifications}
	
	\renewcommand{\arraystretch}{1.25}
	
	\begin{tabularx}{\textwidth}{@{}
			p{2.8cm}
			p{3.2cm}
			>{\raggedright\arraybackslash}X
			@{}}
		\toprule
		
		\textbf{Classifier} &
		\textbf{Preprocessing} &
		\textbf{Model specification} \\
		
		\midrule
		
		Support Vector Machine
		&
		Median imputation;
		standardisation
		&
		Linear SVM (\texttt{LinearSVC});
		$C=1.0$; maximum 10,000 iterations.
		\\
		
		\addlinespace[5pt]
		
		Random Forest
		&
		Median imputation
		&
		500 decision trees; square-root feature sampling
		at each split; all available processor cores
		used during training.
		\\
		
		\addlinespace[5pt]
		
		Neural Network
		&
		Median imputation;
		standardisation
		&
		Feed-forward network with three hidden layers
		containing 128, 64 and 32 units, respectively
		(ReLU activation). Dropout of 0.20 after the
		first two hidden layers. Single sigmoid output
		unit; Adam optimiser; binary cross-entropy loss.
		\\
		
		\bottomrule
		
	\end{tabularx}
	
	\par\vspace{6pt}
	
	\begin{minipage}{\textwidth}
		\footnotesize
		\textit{Note:}
		All classifiers were evaluated using 10-fold
		cross-validation. Training data within each fold
		were used to estimate preprocessing parameters.
		The results represent the combined predictions
		from the 10 held-out folds.
	\end{minipage}
	
\end{table}

	The groups of features used in the following analysis are:
	
	\begin{itemize}
		\item Structural - These are the 18 variables calculated from the letters of words. They include LENGTH, SYLLABLES etc. Note that the number of syllables is taken from the original data source (\cite{ferrand2010french}) because the tested methods of calculating syllables from French words were not efficient. 
			
		\item LENGTH - Number of letters. This is the same variable extracted from the 18 structural variables. However, we examine word length as a single predictor of complexity and in combination with LPP variables because complexity is highly associated with length. 
		
		\item First 6 LPP positions (n = 202) - These are the first 6 LPP positions calculated as above. Note that these include both the variables that were found to be significant after Bonferroni correction and those that were not. Potentially, this would give us a set of variables = 42 × 6 = 252. However, some of these did not have data and so were removed giving a final figure of 202. 
		
		\item All LPP Positions - This is the full cohort of LPP variables. There are potentially 1050 ( 42 × 25) variables. However, the variables with no data were removed from the matrix of results giving a final figure of 441.

	\end{itemize}
	\subsection{Classifier Results}

Table \ref{tab:classification-accuracy} shows the results from the classifier experiments

\begin{table}[H]
	\centering
	\caption{Classification accuracy details}
	\label{tab:classification-accuracy}
	
	\begin{tabular}{@{}lccc@{}}
		\toprule
		\textbf{Feature Set} & 
		\textbf{NN} & 
		\textbf{SVM} & 
		\textbf{RF} \\
		\midrule
		
		Length                         & 0.7727 & 0.7727 & 0.7727 \\
		Structural                     & 0.7785 & 0.7775 & 0.7620 \\
		Structural + All Positions     & 0.8111 & 0.8031 & 0.8085 \\
		First 6 Positions              & 0.7657 & 0.6932 & 0.8012 \\
		All Positions                  & 0.8091 & 0.7966 & 0.8364 \\
		Structural + First 6 Positions & 0.8184 & 0.7951 & 0.8031 \\
		First 6 Positions + Length     & 0.8173 & 0.7859 & \textbf{0.8395} \\
		
		\bottomrule
	\end{tabular}
\end{table}

	The results show that Length alone accounts for 77.27\% of correct classifications.  All three classifiers score the same with this one variable, suggesting a high degree of monotonicity. All classifiers have the same distribution of errors, with sensitivity = 68.30\% and specificity = 86.24%
	
	The Structural variables, which include Length, attain a classification accuracy of 77.85\% with the neural network. Given the small improvement from Length alone, the structural variables other than length add very little to the signal.
	
	The Structural and All Positions features boost these results to 81.11\%, with the Neural Network, indicating that the positional data carry significant information. The Random Forest results are lower than the RF result for All Positions features without the Structural features, indicating that there may be some degradation in classifier accuracy due to high levels of redundancy.
	
	The highest accuracy is achieved by the First 6 Positions + Length in the Random Forest at 83.95\%. This is marginally higher than All Positions on their own at 83.64\%. What this indicates is that there is inherent length data in the All Positions data (due to the mere presence of positional data at positions > 6), but the purely orthographic data in the First 6 positions carries a significant signal. When we add the Length variable to the First 6 Positions features we include the important orthographical information as well as the length information.

	\section{Discussion}
	\label{sec:discussion}
	
	We have seen that there are LPPs that are highly statistically associated with complexity. We have also seen that there are patterns of vowels, consonants and articulation phenomena that are associated with complexity. The indication is that there is information encoded in orthography that is beyond mere words structural information. Importantly, this information can be used to determine the simplicity/complexity of words. A classifier with LPPs from the first 6 positions along with word length was found to be the most predictive of word complexity out of a number of feature sets, including one consisting of all positions. 
	
	We can draw from these observations that there is something in orthography that carries a signal associated with complexity. Certainly length also carries a signal, but, as we have seen, combining length with positional data provides the most robust signal. 
	
	In reviewing the work on phonotactic probability, we noted above that phonotactic probability is an important influence on the way words are processed, both in producing speech and understanding written and spoken speech. In the current study we posit that letter positional probabilities, rather than phonotactic probabilities, carry information about word complexity. 
	
	We also noted that position is important. In \cite{grainger1993masked} above, we noted similarly that speakers of a language are sensitive to the frequencies at which letters occur at different positions. Thus, given a prime of \%BCDE, and a prime of A\%CDE and a target of ABCDE, the first prime will be more efficient than the second if 'A' in first position is less frequent in the language and 'B' in second position. Again, this overlaps with the current study.
	
	The general approach of the majority of discussed studies is that native language users are sensitive to the atomic structure of words. The current study does not disagree with this, but the fundamental question is about word complexity. In the current study, the two extremes of the complexity word sample were chosen on the basis of a lexical decision experiment, and this means that the sample instantiates a phenomenon about word recognition that is similar to the other probability-based studies. The major difference is that the current study, after creating the sample based on lexical decisions, asks what it is about the structure of words at the letter level that distinguishes between simple and complex words. The latter use the structure of words to explain lexical decision criteria. In short, the process of the current study and the other studies is reversed. This is the fundamental difference between the two approaches. However, none of these, including the current study, directly address what is possibly the more interesting question: given that the structure of words carries significant information, what explains how words evolved to have such a structure? The question here is not whether complex words have a different structure compared to simple words, as we have already established this. The question is why this should be so. Although a full analysis of this is beyond the scope of this paper, this is an important issue to address.  
	
	One possibility is that words change over time and their orthography follows. A possible method for this to take place was given by Zipf \cite{zipf1929relative}. He found that pronunciation changes over time according to how frequently the word is used. The idea here is that we look for ways to reduce our effort in communication, so words that are used more often are likely to suffer simplification because there they will be recognised without highly complex articulation forms because they are so common. The argument from Zipf is that 1) a given locution is used often, 2) the phonemes are repeated often due to frequent use 3), 'shortcuts' in articulation take place, but because of the frequency of use, other speakers of the language still understand the locution 4), phonetic 'weakening' takes place,  5) the weakening becomes conventional 6) the lexical form of the original form is altered. Thus, more frequent terms become more easy to pronounce over time.
	
	There is one seemingly contradictory finding from \cite{pagel2007frequency} found that across 200 word meanings from 87 Indo-European languages, common words were less likely to change. However, this is consistent with Zipf's argument in that we might consider that the simplified words, once having achieved their simplified forms, represent collections of phonemes that are the easiest to articulate in the language. There is no behavioural reason for changing them from this form.  
	
	The implication here is that, because the simplification process took place over time, the now simple words had a more complex phonological structure at some time in the past. The words that were complex in the past and remain complex now represent the default level of complexity, and those that have become simplified represent the changed words a simplified new phonology. This suggests that the significant LPP variables are picking up on a change that has occurred over time. It is not a case that there is was initially a structural difference between simple and complex words, but it is true that the structural difference emerged over time. 
	
	Hypothetically, the reason that there are significant LPPs is that the simplification process is not arbitrary. The words that are more common change in a way that is systematic for a given language. As \cite{mahowald2018word} found, words that have a highly probable (more likely) form within the language are more frequent, as are words with more orthographic neighbours.
	
	
	Finally, we need to consider the possibility that the concept of complexity we have been using is actually frequency. As noted above, frequency is often used as a proxy for complexity. We have used a sample consisting of what we have taken to be high and low complexity words. However it is most likely that these also represent the extremes of the frequency range. It is theoretically possible to test for frequency effects by including in the modeling procedure a measure of frequency (although choosing what corpus to base this on might be problematic). Again, this is a possible avenue of enquiry in subsequent work.

	\section{Conclusion}
	
	This study has examined the role of LPPs in carrying a simplicity/complexity signal in French words. We examined the distribution of LPPs in simple as opposed to complex words. We found that there were 145 LPPs that were significant out of a potential cohort of 441. Because the average length of simple words was 6.69, we focused on the first 6 positions because we wished to reduce the confounding effect of length of LPPs in positions >6. We also found that several structural variables were associated with complexity, with length being particularly predictive of complexity in our sample.
	
	We found that there were a number of phenomena associated with LPPs in relation to complexity. We found that 5 out of 6 of the significant LPPs containing an accented letter were associated with lower complexity, while 6 out of 6 of the significant LPPs containing an unaccented letter were associated with higher complexity. The overall finding was that there are significant differences in the orthography of simple as opposed to complex words.
	

	We went on to create classifiers using LPPs and structural data as features to predict binary simple and complex words. We found that length alone achieves an accuracy of 77.27\%. Additional structural variables added little to classification accuracy. The best classifier achieved an accuracy of 83.95\% and consisted of the first 6 LPP features plus Length. We concluded from this that both orthography and length are significant features associated with complexity. 
	
	The question we went on to examine is how orthography can encode complexity. Certainly it is conventionally accepted that word length is associated with complexity, and in the current study it was indeed the variable most closely associated with complexity. However, positing the idea that the placement of individual letters is predictive of complexity is significantly more controversial. 
	
	We considered an explanation from Zipf that more common words become simplified over time due to articulatory economy. The important point here for our purposes is that simplicity is associated with frequency in that words that are repeated often become simplified.

	
	We observed that the question of the association between vowel/consonant patterns and complexity may be worth considering in subsequent work, as is the possibility that there may be a difficulty in disentangling the difference between word complexity and word frequency.


	
	\section*{Funding}
	
	This research received no external funding.

	
	\section*{Conflict of Interest}
	
	The author declares no conflict of interest.

	
	\section*{Data Availability}
	
	The Python scripts used to generate the data used in this study can be obtained from the author at admin@michaeldalvean.com.

	\printbibliography
	\newpage
	
\appendix

\renewcommand{\thesection}{Appendix \Alph{section}}
\section{Low- and High-Complexity Means for Significant Variables}
\label{app:high-low-frequency}

\begin{longtable}{@{}>{\raggedright\arraybackslash}p{0.32\textwidth}
		>{\centering\arraybackslash}p{0.18\textwidth}
		>{\centering\arraybackslash}p{0.18\textwidth}
		>{\centering\arraybackslash}p{0.20\textwidth}@{}}
	
	\caption{Mean values for low- and high-complexity French words and Bonferroni-adjusted $p$-values.}
	\label{tab:french-low-high-complexity}\\
	
	\toprule
	\textbf{Variable}
	&
	\shortstack{\textbf{Mean -- Low}\\\textbf{Complexity}}
	&
	\shortstack{\textbf{Mean -- High}\\\textbf{Complexity}}
	&
	\textbf{$p$ value}
	\\
	
	&
	\textbf{($n=5{,}000$)}
	&
	\textbf{($n=5{,}000$)}
	&
	\textbf{(Bonf. $\times 459$)}
	\\
	\midrule
	\endfirsthead
	
	\multicolumn{4}{c}{\tablename\ \thetable{} -- continued from previous page} \\
	
	\toprule
	\textbf{Variable}
	&
	\shortstack{\textbf{Mean -- Low}\\\textbf{Complexity}}
	&
	\shortstack{\textbf{Mean -- High}\\\textbf{Complexity}}
	&
	\textbf{$p$ value}
	\\
	
	&
	\textbf{($n=5{,}000$)}
	&
	\textbf{($n=5{,}000$)}
	&
	\textbf{(Bonf. $\times 459$)}
	\\
	\midrule
	\endhead
	
	\midrule
	\multicolumn{4}{r}{Continued on next page} \\
	\endfoot
	
	\bottomrule
	\endlastfoot
	
	LENGTH & 6.694 & 9.766 & $<.0001$ \\
	SYLLABLES & 2.071 & 3.050 & $<.0001$ \\
	UNIQUE\_\allowbreak LETTERS & 5.922 & 7.562 & $<.0001$ \\
	LETTER\_\allowbreak DIVERSITY & 0.897 & 0.790 & $<.0001$ \\
	REPEATED\_\allowbreak LETTERS & 0.773 & 2.204 & $<.0001$ \\
	VOWELS & 2.955 & 4.284 & $<.0001$ \\
	CONSONANTS & 3.740 & 5.481 & $<.0001$ \\
	CV\_\allowbreak TRANSITIONS & 4.138 & 6.339 & $<.0001$ \\
	LETTER\_\allowbreak ENTROPY & 2.485 & 2.796 & $<.0001$ \\
	REPETITION\_\allowbreak RATIO & 0.103 & 0.210 & $<.0001$ \\
	ORTH\_\allowbreak NEIGHBOURS & 0.697 & 0.112 & $<.0001$ \\
	CONSONANT\_\allowbreak CLUSTERS & 0.888 & 1.470 & $<.0001$ \\
	e10 & 0.009 & 0.102 & $<.0001$ \\
	MAX\_\allowbreak CONSONANT\_\allowbreak CLUSTER & 1.761 & 1.990 & $<.0001$ \\
	i9 & 0.003 & 0.082 & $<.0001$ \\
	i8 & 0.012 & 0.098 & $<.0001$ \\
	e11 & 0.003 & 0.074 & $<.0001$ \\
	a8 & 0.007 & 0.081 & $<.0001$ \\
	i7 & 0.027 & 0.120 & $<.0001$ \\
	i10 & 0.001 & 0.058 & $<.0001$ \\
	a9 & 0.002 & 0.061 & $<.0001$ \\
	e9 & 0.035 & 0.124 & $<.0001$ \\
	DOUBLE\_\allowbreak LETTERS & 0.157 & 0.308 & $<.0001$ \\
	t10 & 0.007 & 0.071 & $<.0001$ \\
	BIGRAM\_\allowbreak DIVERSITY & 0.995 & 0.982 & $<.0001$ \\
	e12 & 0.000 & 0.051 & $<.0001$ \\
	n11 & 0.002 & 0.056 & $<.0001$ \\
	s10 & 0.009 & 0.069 & $<.0001$ \\
	t12 & 0.001 & 0.046 & $<.0001$ \\
	s11 & 0.003 & 0.050 & $<.0001$ \\
	a7 & 0.023 & 0.091 & $<.0001$ \\
	t11 & 0.002 & 0.047 & $<.0001$ \\
	n10 & 0.005 & 0.054 & $<.0001$ \\
	n9 & 0.013 & 0.069 & $<.0001$ \\
	i11 & 0.000 & 0.039 & $<.0001$ \\
	s12 & 0.000 & 0.037 & $<.0001$ \\
	a10 & 0.000 & 0.036 & $<.0001$ \\
	MEAN\_\allowbreak SYLLABLE\_\allowbreak LENGTH & 3.560 & 3.293 & $<.0001$ \\
	e13 & 0.000 & 0.033 & $<.0001$ \\
	t9 & 0.020 & 0.071 & $<.0001$ \\
	n12 & 0.001 & 0.031 & $<.0001$ \\
	t8 & 0.030 & 0.085 & $<.0001$ \\
	n8 & 0.024 & 0.073 & $<.0001$ \\
	o6 & 0.011 & 0.051 & $<.0001$ \\
	s13 & 0.000 & 0.025 & $<.0001$ \\
	e5 & 0.168 & 0.092 & $<.0001$ \\
	o7 & 0.006 & 0.039 & $<.0001$ \\
	l8 & 0.008 & 0.042 & $<.0001$ \\
	n13 & 0.000 & 0.023 & $<.0001$ \\
	l9 & 0.002 & 0.029 & $<.0001$ \\
	n2 & 0.035 & 0.086 & $<.0001$ \\
	s9 & 0.027 & 0.072 & $<.0001$ \\
	a6 & 0.042 & 0.094 & $<.0001$ \\
	t13 & 0.001 & 0.023 & $<.0001$ \\
	é6 & 0.056 & 0.017 & $<.0001$ \\
	i1 & 0.020 & 0.060 & $<.0001$ \\
	r10 & 0.005 & 0.031 & $<.0001$ \\
	a11 & 0.000 & 0.019 & $<.0001$ \\
	i6 & 0.053 & 0.105 & $<.0001$ \\
	e6 & 0.161 & 0.097 & $<.0001$ \\
	r9 & 0.013 & 0.045 & $<.0001$ \\
	i12 & 0.000 & 0.017 & $<.0001$ \\
	t14 & 0.000 & 0.017 & $<.0001$ \\
	e14 & 0.000 & 0.017 & $<.0001$ \\
	o2 & 0.195 & 0.129 & $<.0001$ \\
	n7 & 0.037 & 0.078 & $<.0001$ \\
	o10 & 0.001 & 0.019 & $<.0001$ \\
	m8 & 0.001 & 0.018 & $<.0001$ \\
	r11 & 0.001 & 0.017 & $<.0001$ \\
	m9 & 0.001 & 0.017 & $<.0001$ \\
	l7 & 0.018 & 0.047 & $<.0001$ \\
	o5 & 0.040 & 0.078 & $<.0001$ \\
	e8 & 0.082 & 0.132 & $<.0001$ \\
	o8 & 0.006 & 0.027 & $<.0001$ \\
	l11 & 0.000 & 0.013 & $<.0001$ \\
	s14 & 0.000 & 0.012 & $<.0001$ \\
	l10 & 0.001 & 0.014 & $<.0001$ \\
	d8 & 0.001 & 0.015 & $<.0001$ \\
	o9 & 0.003 & 0.018 & $<.0001$ \\
	u10 & 0.001 & 0.014 & $<.0001$ \\
	e15 & 0.000 & 0.010 & $<.0001$ \\
	u9 & 0.004 & 0.019 & $<.0001$ \\
	r8 & 0.033 & 0.063 & $<.0001$ \\
	o11 & 0.000 & 0.011 & $<.0001$ \\
	p6 & 0.004 & 0.020 & $<.0001$ \\
	m11 & 0.000 & 0.010 & $<.0001$ \\
	n14 & 0.000 & 0.010 & $<.0001$ \\
	m10 & 0.001 & 0.011 & $<.0001$ \\
	i13 & 0.000 & 0.009 & $<.0001$ \\
	r12 & 0.000 & 0.009 & $<.0001$ \\
	j1 & 0.019 & 0.005 & $<.0001$ \\
	a12 & 0.000 & 0.008 & $<.0001$ \\
	v1 & 0.044 & 0.021 & $<.0001$ \\
	o12 & 0.000 & 0.008 & $<.0001$ \\
	TRIGRAM\_\allowbreak DIVERSITY & 1.000 & 0.998 & $<.0001$ \\
	u12 & 0.000 & 0.008 & $<.0001$ \\
	l6 & 0.028 & 0.052 & $<.0001$ \\
	d7 & 0.005 & 0.018 & $<.0001$ \\
	é9 & 0.006 & 0.019 & $<.0001$ \\
	i5 & 0.089 & 0.127 & $<.0001$ \\
	é10 & 0.000 & 0.009 & $<.0001$ \\
	l12 & 0.000 & 0.008 & $<.0001$ \\
	z6 & 0.011 & 0.002 & $<.0001$ \\
	l2 & 0.043 & 0.022 & $<.0001$ \\
	u11 & 0.000 & 0.008 & $<.0001$ \\
	VOWEL\_\allowbreak CONSONANT\_\allowbreak RATIO & 0.872 & 0.828 & $<.0001$ \\
	c8 & 0.002 & 0.012 & $<.0001$ \\
	z7 & 0.013 & 0.002 & $<.0001$ \\
	a4 & 0.069 & 0.102 & $<.0001$ \\
	s15 & 0.000 & 0.007 & $<.0001$ \\
	t7 & 0.050 & 0.079 & $<.0001$ \\
	c7 & 0.007 & 0.021 & $<.0001$ \\
	t15 & 0.000 & 0.007 & $<.0001$ \\
	m6 & 0.014 & 0.031 & $<.0001$ \\
	m7 & 0.007 & 0.021 & $<.0001$ \\
	ê2 & 0.007 & 0.000 & $<.0001$ \\
	n15 & 0.000 & 0.006 & $<.0001$ \\
	m2 & 0.015 & 0.032 & $<.0001$ \\
	m5 & 0.012 & 0.028 & $<.0001$ \\
	c9 & 0.001 & 0.008 & $<.0001$ \\
	r5 & 0.077 & 0.108 & $<.0001$ \\
	b10 & 0.000 & 0.006 & $<.0001$ \\
	é11 & 0.000 & 0.006 & $<.0001$ \\
	y3 & 0.014 & 0.004 & $<.0001$ \\
	c6 & 0.018 & 0.034 & $<.0001$ \\
	h9 & 0.000 & 0.005 & $<.0001$ \\
	z5 & 0.007 & 0.001 & $<.0001$ \\
	
	p9 & 0.000 & 0.006 & $1.15\times10^{-4}$ \\
	r7 & 0.048 & 0.073 & $1.43\times10^{-4}$ \\
	b8 & 0.001 & 0.007 & $1.49\times10^{-4}$ \\
	c11 & 0.000 & 0.005 & $1.49\times10^{-4}$ \\
	f6 & 0.003 & 0.012 & $1.68\times10^{-4}$ \\
	u3 & 0.087 & 0.060 & $1.88\times10^{-4}$ \\
	r4 & 0.061 & 0.088 & $2.22\times10^{-4}$ \\
	â2 & 0.008 & 0.001 & $2.44\times10^{-4}$ \\
	l1 & 0.032 & 0.017 & $2.59\times10^{-4}$ \\
	é5 & 0.040 & 0.023 & $7.14\times10^{-4}$ \\
	r13 & 0.000 & 0.005 & $7.16\times10^{-4}$ \\
	d9 & 0.001 & 0.007 & $8.10\times10^{-4}$ \\
	h10 & 0.000 & 0.006 & $9.06\times10^{-4}$ \\
	h3 & 0.005 & 0.014 & $1.08\times10^{-3}$ \\
	g7 & 0.003 & 0.011 & $1.12\times10^{-3}$ \\
	é12 & 0.000 & 0.004 & $1.21\times10^{-3}$ \\
	l13 & 0.000 & 0.004 & $1.21\times10^{-3}$ \\
	s8 & 0.052 & 0.074 & $1.23\times10^{-3}$ \\
	a1 & 0.063 & 0.088 & $1.47\times10^{-3}$ \\
	t3 & 0.061 & 0.085 & $1.48\times10^{-3}$ \\
	o4 & 0.060 & 0.084 & $1.48\times10^{-3}$ \\
	u8 & 0.011 & 0.023 & $2.03\times10^{-3}$ \\
	û3 & 0.005 & 0.000 & $2.54\times10^{-3}$ \\
	g6 & 0.014 & 0.027 & $2.72\times10^{-3}$ \\
	f1 & 0.053 & 0.034 & $2.88\times10^{-3}$ \\
	r3 & 0.082 & 0.108 & $3.01\times10^{-3}$ \\
	t16 & 0.000 & 0.004 & $3.45\times10^{-3}$ \\
	t6 & 0.059 & 0.082 & $3.70\times10^{-3}$ \\
	x3 & 0.008 & 0.002 & $4.11\times10^{-3}$ \\
	m12 & 0.000 & 0.004 & $5.31\times10^{-3}$ \\
	â6 & 0.000 & 0.004 & $5.85\times10^{-3}$ \\
	b11 & 0.000 & 0.004 & $5.85\times10^{-3}$ \\
	u13 & 0.000 & 0.004 & $5.85\times10^{-3}$ \\
	g8 & 0.002 & 0.009 & $8.21\times10^{-3}$ \\
	è8 & 0.000 & 0.004 & $8.97\times10^{-3}$ \\
	q11 & 0.000 & 0.004 & $9.91\times10^{-3}$ \\
	
\end{longtable}

\end{document}